\documentclass{elsart}

\usepackage{amssymb}
\newcommand{\maj}{\mbox{MAJ}}

\newtheorem{theorem}{\bf Theorem}
\newtheorem{lemma}{\bf Lemma}
\newtheorem{corollary}{\bf Corollary}

\newtheorem{definition}{\bf Definition}
\newtheorem{remark}{\bf Remark}

\newenvironment{proof}{\par \bf Proof. \rm}{$\Box$ \vspace{1ex}}

\begin{document}
\date{}

\begin{frontmatter}

\title{Sharpening Occam's Razor
}

\author{Ming Li, 
}
\author{John Tromp, 
}
\author{Paul Vit\'{a}nyi
}


\setlength{\textwidth}{6in}
\setlength{\oddsidemargin}{0.0in}
\setlength{\evensidemargin}{0.0in}
\setlength{\textheight}{8in}

%


\begin{abstract}
We provide a new representation-independent 
formulation of Occam's razor theorem, based on
Kolmogorov complexity. This new formulation allows us to:
(i) Obtain better sample complexity than both length-based \cite{blumer1}
and VC-based \cite{blumer} versions of Occam's razor theorem, in many
applications; and (ii)
Achieve a sharper reverse of Occam's razor theorem than that of
\cite{board}. Specifically, we weaken the assumptions
made in \cite{board} and extend the reverse to superpolynomial 
running times.
\end{abstract}

\end{frontmatter}

\section{Introduction} \label{introsec}
Occam's razor theorem as formulated
by \cite{blumer,blumer1} is arguably the substance of efficient pac learning. 
Roughly speaking, it says that in order to (pac-)learn, it suffices to compress.
A partial reverse, showing the necessity of compression,
has been proved by Board and Pitt \cite{board}.
Since the theorem is about the relation between effective
compression and pac learning, it is natural to assume that
a sharper version ensues by couching it in terms
of the {\em ultimate} limit to effective compression which is 
the Kolmogorov complexity. We present results in that direction.

Despite abundant research generated by its importance, 
several aspects of Occam's razor
theorem remain unclear. There are basically two versions.
The {\em VC dimension-based version} of Occam's razor theorem
(Theorem 3.1.1 of \cite{blumer})
gives the following upper bound on sample complexity:
For a hypothesis
space $H$ with $VCdim(H)=d$, $1 \leq d < \infty$,
\begin{equation}\label{vc-sample}
m(H,\delta , \epsilon ) \leq  \frac{4}{\epsilon} 
(d \log \frac{12}{\epsilon} + \log \frac{2}{\delta} ).
\end{equation}
The following lower bound was proved by Ehrenfeucht {\it et al} \cite{ehren}.
\begin{equation}\label{vc-lowerbound}
m(H,\delta , \epsilon ) > \max (\frac{d-1}{32 \epsilon},
\frac{1}{\epsilon} \ln \frac{1}{\delta} ).
\end{equation}
The upper bound in (\ref{vc-sample}) and the lower
bound in (\ref{vc-lowerbound}) differ by a factor
$\Theta (\log \frac{1}{\epsilon} )$. It was shown in
\cite{haussler} that this factor is, in a sense, unavoidable.

When $H$ is finite, one can directly obtain the following bound
on sample complexity for a consistent algorithm:
\begin{equation}\label{direct-sample}
m(H,\delta , \epsilon ) \leq \frac{1}{\epsilon} \ln \frac{|H|}{\delta}.
\end{equation}
For a graded boolean space $H_n$, we have the 
following relationship between
the VC dimension $d$ of $H_n$ and the cardinality of $H_n$,
\begin{equation}
d \leq \log |H_n | \leq nd.
\end{equation}

When $\log |H_n|=O(d)$ holds, then the sample complexity upper bound
given by (\ref{direct-sample}) can be seen to equal
$\frac{1}{\epsilon} (O(d)+\ln \frac{1}{\delta})$ which matches the lower bound
of (\ref{vc-lowerbound}) up to a constant factor,
and thus every consistent
algorithm achieves optimal sample complexity for such hypothesis spaces.

The {\em length-based version} of Occam's razor theorem then
gives the following sample complexity $m$ to guaranty that
the algorithm pac-learns:
For given $\epsilon$ and $\delta$:
\begin{equation}\label{length-sample}
m = \max (\frac{2}{\epsilon} \ln \frac{1}{\delta} , 
(\frac{(2\ln 2)s^{\beta}}{\epsilon} )^{1/(1-\alpha)} ) ,
\end{equation}
This bound is based on  the {\em length-based}
Occam algorithm \cite{blumer}:
A {\em deterministic} algorithm that returns a consistent hypothesis of 
length at most $m^\alpha s^\beta$, where $\alpha < 1$ and $s$ is the length
of the target concept. 


In summary, the VC dimension based Occam's razor theorem 
may be hard to use and it sometimes does not give the best sample
complexity. The length-based Occam's razor is more convenient
to use and often gives better sample complexity in the discrete case.

However, as we demonstrate below, the fact that the length-based
Occam's razor theorem sometimes gives inferior sample
complexity, can be due to the redundant representation format of the concept.
We believe Occam's razor theorem should be
``representation-independent''. That is, it should not be dependent
on  accidents of ``representation format''.  (See \cite{manfred} for 
other representation-independence issues.) In fact, the sample
complexities given in (\ref{vc-sample}) and (\ref{vc-lowerbound})
are indeed representation-independent. However they are not
easy to use and do not give optimal sample complexity.
Here, we give a Kolmogorov complexity based Occam's razor
theorem. We will demonstrate that our KC-based Occam's razor theorem
is convenient to use (as convenient as the length based
version), gives a better sample complexity than the 
length based version, and is representation-independent.
In fact, the length based version
can be considered as a specific computable approximation
to the KC-based Occam's razor.

As one of the examples, we will demonstrate that the standard trivial learning
algorithm for monomials actually often
has a {\it better sample complexity}
than the more sophisticated Haussler's greedy algorithm \cite{hauss}.
This is
contrary to the commen, but mistaken,  
belief that Haussler's algorithm is better
in all cases (to be sure, Haussler's method is superior
for target monimials of small length).
Another issue related to Occam's razor theorem is the
status of the reverse assertion.
Although a partial reverse of Occam's razor theorem has
been proved by \cite{board}, it applied only to the case of
polynomial running time and sample complexity. 
They also required a property
of closure under exception list. This latter requirement, although
quite general, excludes some reasonable concept classes. Our new
formulation of Occam's razor theorem allows us to
prove a more general reverse of Occam's razor
theorem, allowing the arbitrary running time and weakening
the requirement of exception list of \cite{board}.

\footnote{A preliminary version was presented at 
the {\em 8th Intn'l Computing and Combinatorics Conference
(COCOON)}, held in Singapore, August, 2002.
}

{\bf Discussion of Result and Technique:}
In our approach we obtain better 
bounds on the sample complexity to learn the representation of
a target concept in the given representation system. 
These bounds, however, are representation-independent
and depend only on the Kolmogorov complexity of the target concept.
If we don't care about the representation of the hypothesis
(but that is not the case in this paper) then better ``iff Occam style''
characterizations of polynomial time learnability/predicatability
can be given. They rely 
on Schapire's result that ``weak learnability''
equals ``strong learnability'' in polynomial time \cite{Sch90}
exploited in \cite{HeWa95}. For a recent survey of  
the important related ``boosting'' technique see \cite{Sch02}.

The use of Kolmogorov complexity is to obtain a bound on the
size of the hypotheses class for a fixed (but arbitrary)
target concept.
Obviously, the results described
can be obtained using other proof methods---all true provable statements
must be provable from the axioms of mathematics by the inference methods
of mathematics. The question is whether
a particular proof method facilitates and guides the proving effort.
The message we want to convey is that thinking in terms of coding
and incompressibility suggest improvements to long-standing results.
 A survey of the use of the Kolmogorov complexity
method in combinatorics, computational complexity, and
the analysis of algorithms is \cite{lv} Chapter 6.

\section{Occam's Razor}
Let us assume the usual definitions, say Anthony and Biggs \cite{anthony},
and notation of \cite{board}. For
Kolmogorov complexity we assume the basics of \cite{lv}.

In the following $\Sigma, \Gamma$ is are finite {\em alphabets}: We
consider only discrete learning problems in this paper. 
The set of finite strings over $\Sigma$ is denoted by $\Sigma^*$
and similarly for $\Gamma$.
An element of $\Sigma^*$ is an {\em example}, and a {\em concept}
is a set of examples (a language over $\Sigma$). 
An {\em representation} is an element of $\Gamma^*$.

\begin{definition}
A {\em representation system} is a tuple $(R,\Gamma , c , \Sigma )$, where
$R \subset \Gamma^*$ is the set of representations, and
$c:R \rightarrow 2^{\Sigma^*}$ maps representations to concepts, the latter
being languages over $\Sigma$.
\end{definition}

Hence, given $R$ the mapping $c$ determines a {\em concept class}.
For example, let $\Gamma$ is the alphabet to express Boolean formulas,
$\Sigma = \{0,1\}$, and let
$R$ be the subset of disjunctive normal form (DNF) formulas.
Let $c$ map each element $r \in R$, say a DNF formula over $n$
variables, to $c(r) \subseteq \{0,1\}^n$ such that every example
$e \in c(r)$ viewed as truth-value assignment makes $r$ ``true''. 
That is, if $e=e_1 \ldots e_n$ and we assign ``true'' or ``false''
to the $i$th variable in $r$ according to whether $e_i$ equals ``0''
or ``1'' then $r$ becomes ``true''. Each concept in the thus defined
concept class is the set of truth assignments that make a particular
DNF formula ``true''. 

\begin{definition}
A {\em pac-algorithm} for a representation system
${\bf R} = (R,\Gamma , c , \Sigma )$ is a randomized algorithm $L$
such that, for every $s,n\geq 1,\epsilon>0,\delta>0,r \in R^{\leq s}$,
and every probability distribution $D$ on $\Sigma^{\leq n}$,
if $L$ is given $s,n,\epsilon,\delta$ as input 
and has access to an oracle providing examples of $c(r)$ (the concept
represented by $r$) according to $D$,
then $L$,
with probability at least $1-\delta$, outputs a representation $r' \in R$
approximating the target $r$ in the sense that
$D(c(r')\Delta c(r)) \leq \epsilon$.
Here, $\Delta$ denotes the symmetric set difference.
\end{definition}
The acronym ``pac'' coined by Dana Angluin
stands for ``probably approximately correct'' which aptly captures the
requirement the output representation must satisfy according to the definition.
The question of interest in pac-learning is how many examples
(and running time) a learning algorithm has to qualify as a pac-alpgorithm. 
The {\em running time} and and number of examples ({\em sample complexity})
of the pac-algorithm are
expressed as functions $t(n,s,\epsilon,\delta)$ and
$m(n,s,\epsilon,\delta)$. The following definition generalizes the
notion of Occam algorithm in \cite{blumer}:

\begin{definition}
\label{def.kcoccam}
An {\em Occam-algorithm} for a representation system
${\bf R} = (R,\Gamma , c , \Sigma )$ is a randomized algorithm which
 for every $s,n\geq 1,  \gamma >0$,
on input of a sample consisting of
$m$ examples of a fixed target $r\in R^{\leq s}$, 
with probability at least $1-\gamma$ outputs a representation $r' \in R$
consistent with the sample, such that $K(r' \mid r,n,s) < m/f(m,n,s,\gamma)$,
with $f(m,n,s,\gamma)$, the compression achieved, being an increasing
function of $m$.
\end{definition}
The {\em length-based version} of (possibly randomized)
 Occam algorithm can be obtained
by replacing $K(r' \mid r,n,s)$ by $|r|$ in this definition.
The {\em running time} of the Occam-algorithm is expressed as a function
$t(m,n,s,\gamma)$, where $n$ is the maximum length of the input examples.

\begin{remark}\label{rem.kco}
\rm
An Occam algorithm satisfying a given $f$, 
achieves a lower bound on the number $m$ of examples required
in terms of $K(r' \mid r,n,s)$, the Kolmogorov complexity of
the outputted representation conditioned on the target representation,
rather than the (maximal) length $s$ of $r$ as in the original Occam
algorithm \cite{blumer} and the length-based version above. 
This improvement enables one to use information drawn from the hidden
target for reduction of the Kolmogorov complexity of the output representation,
and hence further reduction of the required sample complexity.
\end{remark}
We need to show that the main properties
of an Occam algorithm are preserved under this generalization.
Our first theorem is a Kolmogorov complexity based Occam's Razor.
We denote the minimum $m$ such that $f(m,n,s,\gamma) \geq x$ by
$f^{-1}(x,n,s,\gamma)$, where we set $f^{-1}(x,n,s,\gamma)=\infty$ 
if $f(m,n,s,\gamma) < x$
for every $m$.

\begin{theorem}
\label{KCoccam}
Suppose we have an Occam-algorithm for
${\bf R} = (R,\Gamma , c , \Sigma )$ with compression $f(m,n,s,\gamma)$.
Then there is a pac-learning algorithm
for {\bf R} with sample complexity 
\[ m(n,s,\epsilon,\delta) =
   \max \left\{\frac{2}{\epsilon}\ln \frac{2}{\delta},
        f^{-1}(\frac{2\ln 2}{\epsilon},n,s,\delta/2) \right\}, \]
and running time $t_{\mbox{pac}}(n,s,\epsilon,\delta) =
t_{\mbox{occam}}(m(n,s,\epsilon,\delta),n,s,\delta/2)$.
\end{theorem}

\begin{proof}
On input of $\epsilon,\delta,s,n$, the learning algorithm will take a sample
of length $m=m(n,s,\epsilon,\delta)$ from the oracle, then
use the Occam algorithm with $\gamma=\delta/2$ to find a hypothesis
(with probability at least $1-\delta/2$) consistent with the sample and
with low Kolmogorov complexity.
In the proof we abbreviate $f(m,n,s,\gamma)$ to $f(m)$
with the other parameters implicit.
Learnability follows in the standard manner from bounding
(by the remaining $\delta/2$) the probability
that all $m$ examples of the target concept
fall outside the, probability $\epsilon$ or greater,
symmetric difference with a bad hypothesis.
Let $m =  m(n,s,\epsilon,\delta)$. Then 
$m \geq f^{-1} (\frac{2 \ln 2}{\epsilon} ,n,s, \frac{\delta}{2})$ gives
\[ \epsilon - \frac{\ln 2}{f(m)} \geq \frac{\epsilon}{2}, \]
and therefore $m \geq \frac{2}{\epsilon}\ln \frac{2}{\delta}$ gives
\[ m(\epsilon - \frac{\ln 2}{f(m)} ) \geq \ln \frac{2}{\delta} .\]
This implies (taking the exponent on both sides and 
using $1-\epsilon<e^{-\epsilon }$)
\[ 2^{m/f(m)}(1-\epsilon)^{m} \leq \delta/2 .\]
The probability that some concept the Occam-algorithm can output
has all $m$ examples being bad is at most
the number of concepts of complexity less than $m/f(m)$, times
$(1-\epsilon)^m$, which by the above is at most $\delta/2$.
\end{proof}

\begin{corollary}
When the compression is of the form
\[f(m,n,s,\gamma) = \frac{m^{1-\alpha}}{p(n,s,\gamma)},\]
one can achieve a sample complexity of
\[\max\left\{\frac{2}{\epsilon}\ln \frac{2}{\delta},
\left( \frac{(2 \ln 2)p(n,s,\delta/2)}{\epsilon} \right)^{1/(1-\alpha)}\right\}.\]
In the special case of total compression, where $\alpha=0$, this
further reduces to
\begin{equation} \label{total-compression}
\frac{2}{\epsilon}\left\{\max(\ln \frac{2}{\delta},(\ln 2) 
p(n,s,\delta/2))\right\}.
\end{equation}
For deterministic Occam-algorithms, we can furthermore replace
$2/\delta$ and $\delta/2$ in Theorem~\ref{KCoccam} by $1/\delta$ and
$\delta$ respectively.
\end{corollary}

\begin{remark}
\rm
Essentially, our new 
Kolmogorov complexity condition is a computationally
universal generalization of the length condition in the original
Occam's razor theorem of \cite{blumer1}. Here, in
Theorem~\ref{KCoccam}, we consider the 
shortest description length over all effective representations,
given the target representation,
rather than in a specific (syntactical) representation system.
This allows us to bound the required sample complexity
not by a function of the number of hypotheses 
(returned representations) 
of length at most the bound on
the length of the target representation, but by a similar
function of the number of hypotheses
that have a certain Kolmogorov complexity conditioned
on the target concept, see Remark~\ref{rem.kco}.
Nonetheless, like in the original Occam's razor Theorem of \cite{blumer1},
we return a representation of a concept approximating the target 
concept in the given representation  system, rather than
a representation outside the system like in Boosting approaches. 
\end{remark}

Suppose we have a concept $c$ and a mis-classified example $x$---an
{\em exception}. Then, the symmetric difference $c \Delta \{x\}$
classifies $x$ correctly: if $x \not\in c$ then  
$c \Delta \{x\} = c \bigcup  \{x\}$, and if $x \in c$ then
$c \Delta \{x\} = c \setminus  \{x\}$. 

\begin{definition}
An {\em exception handler} for a representation system
${\bf R} = (R,\Gamma , c , \Sigma )$ is an algorithm which
on input of a representation $r\in R$ of length $s$,
and an $x \in \Sigma^{\ast}$ of length $n$,
outputs a representation $r' \in R$ of the concept $c(r) \Delta \{x\}$,
of length at most $e(s,n)$, where $e$ is called the {\em exception expansion}
function.
The running time of the exception-handler is expressed as a function
$t(n,s)$ of the representation and exception lengths.
If $t(n,s)$ is polynomial in $n,s$, and furthermore $e(s,n)$ is of the form
$s+p(n)$ for some polynomial $p$, then we say ${\bf R}$ is {\em polynomially
closed under exceptions}.
\end{definition}


\begin{theorem}
\label{sex}
Let $L$ be a deterministic pac-algorithm
with $m(n,s,\frac{1}{2n},\gamma)$ the sample size,
and let $E$ be an exception handler for
a representation system ${\bf R}$.
Then there is an Occam algorithm for ${\bf R}$
that for $m$ examples achieves compression 
$f(m,n,s,\gamma)= \frac{1}{2\epsilon n}$.
Moreover, $m \geq 2nm(n,s,\frac{1}{2n},\gamma)$ and
where $\epsilon$, depending on $m,n,s,\gamma$, is such that
$m(n,s,\epsilon,\gamma)=\epsilon m$ holds.
\end{theorem}

\begin{proof}
The proof is obtained in a fashion similar to 
\cite{board}.
Suppose we are given a sample of length $m$ and confidence parameter
$\gamma$.
Assume without loss of generality that the sample
contains $m$ different examples.
Define a uniform distribution on these examples with $\mu (x) = 1/m$
for each $x$ in the sample.
Let $\epsilon$ be as described.
The function $m(n,s,\epsilon,\gamma)$ decreases with  
increasing $\epsilon$,
while the function $\epsilon m$ increases with $\epsilon$
so the two necessarily intersect, under the assumption in the theorem,
for some $\epsilon_0$, although it may yield an
$\epsilon_0 >\frac{1}{2n}$, giving no actual compression.
For example, if $m(n,s,\epsilon,\gamma)= (\frac{1}{\epsilon})^{b}$
for some constant $b$, then $\epsilon_0 = m^{-1/(b+1)}$.
Apply $L$ with $\delta = \gamma$ and $\epsilon = \epsilon_0$.
With probability $1-\gamma$,
it produces a concept which is correct with error $\epsilon$,
giving up to $\epsilon m$ exceptions.
We can just add these one by one using
the exception handler.
This will expand the concept size, but not the Kolmogorov complexity.
The resulting representation can be described by the $\leq \epsilon m$
examples used plus the $\leq \epsilon m$ exceptions found,
Since $L$ is deterministic, this uniquely determines the required
consistent concept.
The compression achieved is $\frac{m}{2\epsilon mn} = \frac{1}{2\epsilon n}$.
This is an increasing function of $m$, since increasing the slope of
the function $\epsilon m$ moves its intersection with
the function $m(n,s,\epsilon,\gamma)$ to the left, that is,
to smaller $\epsilon$.
\end{proof}

\begin{definition}
Let ${\bf R} = (R,\Gamma , c , \Sigma )$ be a representation system.
The concept $\maj(r_1,r_2,r_3)$
is the set $\{x :$ $x$ belongs to at least two out of 
the three concepts $c(r_1),c(r_2),c(r_3)\}$.
A {\em majority-of-three algorithm} for
${\bf R}$ is an algorithm which
on input of three representation $r_1,r_2,r_3 \in R^{\leq s}$,
outputs a representation $r' \in R$ of the concept $\maj(r_1,r_2,r_3)$
of length at most $e(s)$, where $e$ is called
 the {\em majority expansion}
function.
The running time of the algorithm is expressed as a function
$t(s)$ of the maximum representation length.
If $t(s)$ and $e(s)$ are polynomial in $s$
then we say ${\bf R}$ is {\em polynomially
closed under majority-of-three}.
\end{definition}

\begin{theorem}
\label{maj}
Let $L$ be a deterministic pac-algorithm with sample complexity
$m(n,s,\epsilon,\delta) \in o(1/\epsilon^2)$, and let $M$
be a majority-of-three algorithm for
the representation system ${\bf R}$.
Then there is an Occam algorithm for ${\bf R}$ that for $m$ examples
has compression $f(m,n,s,\gamma)=m/3nm(n,s,\frac{1}{2\sqrt{m}},\gamma/3)$.
\end{theorem}

\begin{proof}
Let us be given a sample of length $m$.
Take $\delta = \gamma / 3$ and $\epsilon = \frac{1}{2\sqrt{m}}$.

{\it Stage 1:} Define a uniform distribution on the $m$ examples
with $\mu_1 (x) = 1/m$ for each $x$ in the sample.
Apply the learning algorithm.
It produces (with probability at least $1-\gamma/3$)
a hypothesis $r_1$ which has error less than $\epsilon$,
giving up to $\epsilon m = \sqrt{m}/2$ exceptions.
Denote this set of exceptions by $E_1$.

{\it Stage 2:} Define a new distribution
$\mu_2(x) = \epsilon$ for each $x \in E_1$,
and $\mu_2(x) = (1-|E_1|/2\sqrt{m})/(m-|E_1|)$ for each $x \not\in E_1$.
Apply the learning algorithm.
It produces (with probability at least $1-\gamma/3$)
a hypothesis $r_2$ which is correct on all of $E_1$ and with error
less than $\epsilon$ on the remaining examples.
This gives up to $\epsilon (m-|E_1|) / (1-|E_1|/2\sqrt{m}) < \sqrt{m}$
exceptions. This set, denoted $E_2$, is disjoint from $E_1$.

{\it Stage 3:} Define a new distribution on the $m$ examples
with $\mu(x) = 1/|E_1 \cup E_2| > \epsilon$ for each $x$ in $E_1\cup E_2$,
and $\mu(x) = 0$ elsewhere.
Apply the learning algorithm.
The algorithm produces (with probability at least $1-\gamma/3$)
a hypothesis $r_3$ which is correct on all of $E_1$ and $E_2$.

In total the number of examples consumed by the pac-algorithm
is at most $3m(n,s,\frac{1}{2\sqrt{m}},\gamma/3)$, each requiring
$n$ bits to describe.
The three representations are combined into one representation
by the majority-of-three algorithm $M$. This is necessarily correct on all
of the $m$ examples, since the three exception-sets are all disjoint.
Furthermore, it can be described in terms of the
examples fed to the deterministic pac-algorithm
and thus achieves compression
$f(m,n,s,\gamma) = m/3nm(n,s,\frac{1}{2\sqrt{m}},\gamma/3)$.
This is an increasing function of $m$ given the assumed
subquadratic sample complexity. 
\end{proof}

The following corollaries use the fact that if
 a representation system is learnable,
it must have finite VC-dimension and hence,
according to (\ref{vc-sample}), they are learnable with sample
complexity subquadratic in $\frac{1}{\epsilon}$.
\begin{corollary}
Let a representation system ${\bf R}$ be
closed under either exceptions or majority-of-three, or both.
Then ${\bf R}$ is pac-learnable iff
there is an Occam algorithm for ${\bf R}$.
\end{corollary}

\begin{corollary}
Let a representation system ${\bf R}$ be polynomially
closed under either exceptions or majority-of-three, or both.
Then ${\bf R}$ is deterministically polynomially pac-learnable iff
there is a polynomial time Occam algorithm for ${\bf R}$.
\end{corollary}

\noindent
{\it Example.}
Consider threshold circuits,
acyclic circuits whose nodes compute threshold
functions of the form $a_1x_1 + a_2x_2 + \cdots +a_nx_n \geq \delta$,
$x_i \in \{0,1\}, a_i,\delta \in N$ (note that no expressive
power is gained by allowing rational weights and threshold).
A simple way of representing circuits
over the binary alphabet is to number each node and use
{\em prefix-free encodings} of these numbers. For instance, encode $i$
as $1^{|\mbox{bin}(i)|}0\mbox{bin}(i)$,
the binary representation of $i$ preceded by its length in unary.
A complete node encoding then consists of the encoded index, encoded
weights, threshold, encoded degree, and encoded indices of the nodes
corresponding to its inputs. A complete circuit can be encoded with
a node-count followed by a sequence of node-encodings.
For this representation, a majority-of-three
algorithm is easily constructed that renumbers two of its three input
representations, and combines the three by adding a
3-input node computing the majority function
$x_1+x_2+x_3 \geq 2$.
It is clear that under this representation,
the system of threshold circuits
are polynomially closed under majority-of-three.
On the other hand they are not closed under exceptions,
or under the exception lists of \cite{board}.

\noindent
{\it Example.} Let $h_1 , h_2, h_3$ be 3 $k$-DNF formulas.
Then $\maj (h_1,h_2,h_3) = (h_1 \wedge h_2) \vee (h_2 \wedge h_3) \vee
(h_3 \wedge h_1)$ which can be expanded into a $2k$-DNF formula.
This is not good enough for Theorem~\ref{maj}, but it allows us to conclude
that pac-learnability of $k$-DNF implies compression of $k$-DNF into
$2k$-DNF.

\section{Applications}
Our KC-based Occam's razor theorem might
be {\it conveniently} used, providing better sample 
complexity than the length-based version.
In addition to giving better sample complexity, 
our new KC-based Occam's razor theorem,
Theorem~\ref{KCoccam}, is easy to use, as easy
as the length based version, as demonstrated by the following
two examples.
While it is easy to construct an artificial system with
extremely bad representations such that our Theorem~\ref{KCoccam}
gives {\it arbitrarily} better sample complexity than
the length-based sample complexity given in
(\ref{length-sample}), we prefer to give natural examples.

\noindent
{\bf Application 1: Learning a String.}

The DNA sequencing process can be modeled as the problem
of learning a super-long string in the pac model \cite{jiang1,li}.
We are interested in learning a target string $t$ of length $s$,
say $s=3 \times 10^9$ (length of a human DNA sequence).
At each
step, we can obtain as an example a substring of this sequence
of length $n$, from a random location of $t$ (Sanger's Procedure). 
At the time of writing, $n \approx 500$, and 
sampling is very expensive.
Formally, the concepts we are learning are sets of possible length $n$
substrings of a superstring, and these are naturally
represented by the superstrings. We assume a minimal target representation
(which may not hold in practice).
Suppose we obtain a 
sample of $m$ substrings (all positive examples). In biological
labs, a Greedy algorithm which repeatedly merges a pair of substrings
with maximum overlap is routinely used. It is conjectured
that Greedy produces a common superstring $t'$ of length at most $2s$,
where $s$ is the optimal length (NP-hard to find). In \cite{blum}, 
we have shown that $s \leq |t'| \leq 4s$.
Assume that $|t'| \approx 2s$.\footnote{Although only the
$4s$ upper bound was proved in \cite{blum}, which has since been improved,
it is widely believed
that $2s$ is the true bound.} 
Using the length-based Occam's razor theorem, that is, Theorem~\ref{sex}
with $K(r' \mid r,s,n)$ in Definition~\ref{def.kcoccam} replaced
by $|r'|$,
this length of $2s$ would determine the sample complexity,
as in (\ref{total-compression}), with
$p(n,s,\delta/2)= 2 \cdot 2s$
(the extra factor 2 is the 2-logarithm of the size of the alphabet
$\{A,C,G,T\}$).
Is this the best we can do?
It is well-known that the sampling process in DNA sequencing is a very
costly and slow process. 
We improve the sample complexity using our KC-based Occam's razor 
theorem.

\begin{lemma}
Let $t$ be the target string of length $s$ and $t'$ be the 
superstring returned by Greedy of length at most $2s$. Then
\[
K(t' \mid t,s,n ) \leq 2s (2\log s + \log n) / n .
\]
\end{lemma}
\begin{proof}
We give $t'$ a short description using some information
from $t$. Let $S = \{ s_1 , \ldots , s_m \}$ be the set of
$m$ examples (substrings of $t$ of length $n$).
Align these substrings with the common superstring $t'$, from 
left to right. Divide them into groups such that each group's
leftmost string overlaps with every string in the group but
does not overlap with the leftmost string of the previous group. 
Thus there are at most $2s/n$ such groups. 
To specify $t'$, we only need to specify these $2s/n$ groups.
After we obtain the superstring for each group, we re-construct $t'$
by optimally merging the superstrings of neighboring groups.
To specify each group, we only need to specify the first and the last
string of the group and how they are merged. This is because every
other string in the group is a substring of the string obtained by
properly merging the first and last strings. Specifying the first and
the last strings requires $2 \log s$ bits of information
to indicate their locations in $t$ and we need another
$\log n$ bits to indicate how they are merged.
Thus $K(t'\mid t,s,n) \leq 2s (2 \log s + \log n) / n$. 
\end{proof}

This lemma shows that (\ref{total-compression}) can also be
applied with 
$p(n,s,\delta/2)= 2\cdot 2s (2 \log s + \log n) / n$, giving a factor
$n / (2\log s + \log n)$ improvement in sample-complexity.
Note that in (mammal) genome computation practice, 
we have $n=500$ and $s=3 \times 10^9$.
The sample complexity using the Kolmogorov complexity-based
Occam's razor is reduced over the ``length based'' 
Occam's razor by a multiplicative factor of 
$n / (2\log s + \log n) \approx \frac{500}{2 \times 31 + 9} \approx 7$. 

\noindent
{\bf Application 2: Learning a Monomial.}

Consider boolean space of $\{0,1\}^n$. There are two well-known algorithms 
for learning monomials. One is the standard algorithm.

\noindent
{\bf Standard Algorithm.}
\begin{enumerate}
\item
Initially set the concept representation
$M := x_1 \overline{x_1} \ldots x_n \overline{x_n}$
(a conjunction of all literals of $n$ 
variables---which contradicts every example).
\item
For each positive example, delete from the current $M$ the literals that 
contradict the example.
\item
Return the resulting monomial $M$.
\end{enumerate}

Haussler \cite{hauss} proposed a more sophisticated algorithm based
on set-cover approximation as follows.
Let $k$ be the number of variables in the target monomial, and $m$
be the number of examples used.

\noindent
{\bf Haussler's Algorithm.}
\begin{enumerate}
\item
Use only negative examples. 
For each literal $x$, define $S_x$ to be the set of negative examples
such that $x$ falsifies these negative examples.
The sets associated with the literals in the target monomial form a
set cover of negative examples.
\item
Run the approximation algorithm of set cover, this will use at most
$k \log m$ sets or, equivalently, literals in our approximating
monomial. 
\end{enumerate}

It is commonly believed that Haussler's algorithm 
has better sample complexity than the standard algorithm
\footnote{In fact, Haussler's algorithm is specifically aimed
at reducing sample complexity for small target monomials, and that it
does.
}
We demonstrate that the opposite is sometimes true (in fact for
most cases), using our KC-based Occam's razor theorem,
Theorem~\ref{KCoccam}. Assume that our target monomial $M$ is of
length $n - \sqrt{n}$. Then the length-based Occam's razor theorem
gives sample complexity $n/\epsilon$ for both algorithms, by
Formula~\ref{total-compression}. However, 
$K(M' \mid M)\leq \sqrt{n}\log 3+O(1)$,
where $M'$ is the monomial returned by the standard algorithm. This is
true since the standard algorithm always produces a monomial
$M'$ that contains {\em all} literals of the target monomial $M$, and
we need at most $\sqrt{n} \log 3 + O(1)$ bits to specify
whether other literals are in (positive or negative)
or not in $M'$ for the 
variables that are in $M'$ but not in $M$.
Thus our (\ref{total-compression}) gives
the sample complexity of $O(\sqrt{n}/\epsilon)$.
In fact, as long as $|M| > n/\log n$ (which is most likely
to be the case if every monomial has equal probability), 
it makes sense to use the standard algorithm.

\section{Conclusions}

Several new problems are suggested by this work.
If we have an algorithm that, given a length-$m$ sample of a concept
in Euclidean space, produces a consistent hypothesis that can be described
with only $m^\alpha, \alpha<1$ symbols (including a symbol for every real
number; we're using uncountable representation alphabet), then it seems
intuitively appealing that this implies some form of learning.
However, as noted in \cite{board}, 
the standard proof of Occam's Razor does not apply, since we cannot
enumerate these representations. The main open question is under
what conditions (specifically on the real number computation
model) such an implication would nevertheless hold.

Can we replace the exception element or majority of 3 requirement
by some weaker requirement? Or can we even eliminate such
closure requirement and obtain a complete reverse of
Occam's razor theorem?
Our current requirements do not even include things
like k-DNF and some other reasonable representation systems. 

\section{Acknowledgements}
We wish to thank Tao Jiang for many stimulating discussions.

\end{document}